\documentclass[11pt,a4paper]{article}
\usepackage[margin=25mm]{geometry}
\usepackage[T1]{fontenc}
\usepackage{lmodern,amsmath,amssymb,amsthm,booktabs,graphicx,microtype,xcolor,hyperref}
\hypersetup{colorlinks=true,linkcolor=blue!45!black,citecolor=blue!45!black,urlcolor=blue!45!black,pdfauthor={The Viet Bui and Tien Mai},pdftitle={Who Should Verify? Reviewer Selection with Partial Information and Outcome Audits}}
\newtheorem{theorem}{Theorem}
\newtheorem{proposition}[theorem]{Proposition}
\newcommand{\one}{\mathbf{1}}

\title{Who Should Verify?\\Reviewer Selection with Partial Information\\and Outcome Audits}
\author{The Viet Bui and Tien Mai\\Singapore Management University}
\date{Working paper, November 2026}

\makeatletter\let\tableinput\@@input\makeatother
\begin{document}
\maketitle
\begin{abstract}
A team that uses AI reviewers must choose a review procedure: release an analyst's original answer, or release the answer of one of several reviewers who have read it. Making that choice responsibly requires independent evidence, in the form of verified answer keys and sometimes the reviews themselves, and both cost money. We ask how much evidence is needed to certify, on a fixed batch of items, that the chosen procedure gets at most a stated number fewer correct answers than the best procedure on offer, whatever the evidence not yet bought turns out to be. When every review is already available and only keys are bought, an exact stopping rule and an integer program give the minimum number of keys. When reviews and keys both have prices, the two purchases can complement each other, so the usual greedy argument does not apply directly. Given the completed record, a polynomial-time algorithm constructs a certificate costing at most $H_{2(K-1)}=O(\log K)$ times the cheapest certificate in hindsight, for $K$ procedures, any batch size and any prices; unless P equals NP, this logarithmic factor cannot be improved to leading order. Searching for a certificate without hindsight can cost far more: in a family of instances whose cheapest certificate costs one purchase, every valid search policy costs at least $(n+1)/2$ purchases in expectation on some instance, for $n$ unresolved items, and this bound is tight. For two procedures with binary answers, when items are independent and identically distributed under a known law, an optimal search policy depends on five counts and is computed by dynamic programming. We replay saved outputs from three studies: four language models reviewing one another on 2,000 MMLU questions, a small local model checking 896 new numerical reports, and 48 forecast reports on Singapore electricity demand. With reviews already available, the targeted audit used 6.9\% fewer keys than the best fixed order chosen on development data for MMLU and 1.2\% fewer for the numerical reports, where an entropy-based baseline used fewer still. On held-out MMLU questions, the procedure chosen by the targeted audit was 2.94 points more accurate than the reviewer chosen by stand-alone accuracy; choosing with every audit key opened did about as well, and on the numerical reports neither made a difference. When reviews were priced, the constructed certificates were within four price units of the exact minimum in all ten settings, and the policy free to buy either field next matched or slightly improved on the better of two policies that always buy the key first or always buy the review first. All guarantees concern the audited batch; separate experiments measure performance on new items and the absolute quality of the reports.
\end{abstract}

\section{Introduction}
Two language models each answered 803 of the same 1,000 multiple-choice questions correctly on their own. Yet when they reviewed answers drafted by a third model, one finished with 51 more correct answers than the other. Which score should a research group use when choosing who will check its work? In this held-out example from our MMLU study, the group could keep the draft answer from Haiku-5-5 or release a reviewer's answer after the reviewer had read the draft and its justification. Haiku-4-5 corrected 98 of the original errors but changed 122 correct answers into errors, leaving 762 correct. GPT-5.4-mini corrected slightly fewer errors, 93, yet overturned only 66 correct answers, leaving 813 correct. We can see the difference because the archive contains every review and correct answer. A group facing the decision for the first time must work out how much of that evidence it needs to obtain.

To see what the group is buying, consider a single forecast report. The model that drafted it, the \emph{producer}, answers yes when asked whether the mean absolute forecast error exceeded four units. The group can pay a reviewer for a second answer or recompute the statistic from the source data to obtain the verified answer, which we call the \emph{key}. Suppose, for illustration, that a review costs one unit, a key costs four, and the reviewer agrees with the producer three times out of four. If the group opens the review first and sees yes, both procedures release the same answer and earn the same score whatever the data say. Only a no requires the group to buy the key to decide which procedure wins. The expected cost is $1+\tfrac14\times4=2$ units. Buying the key first costs four and also settles the choice immediately. If the original is correct, keep it. If it is wrong, choose the reviewer, who can only tie or improve on it.

\begin{figure}[htbp]
\centering
\includegraphics[width=\linewidth,trim=0 45 0 50,clip]{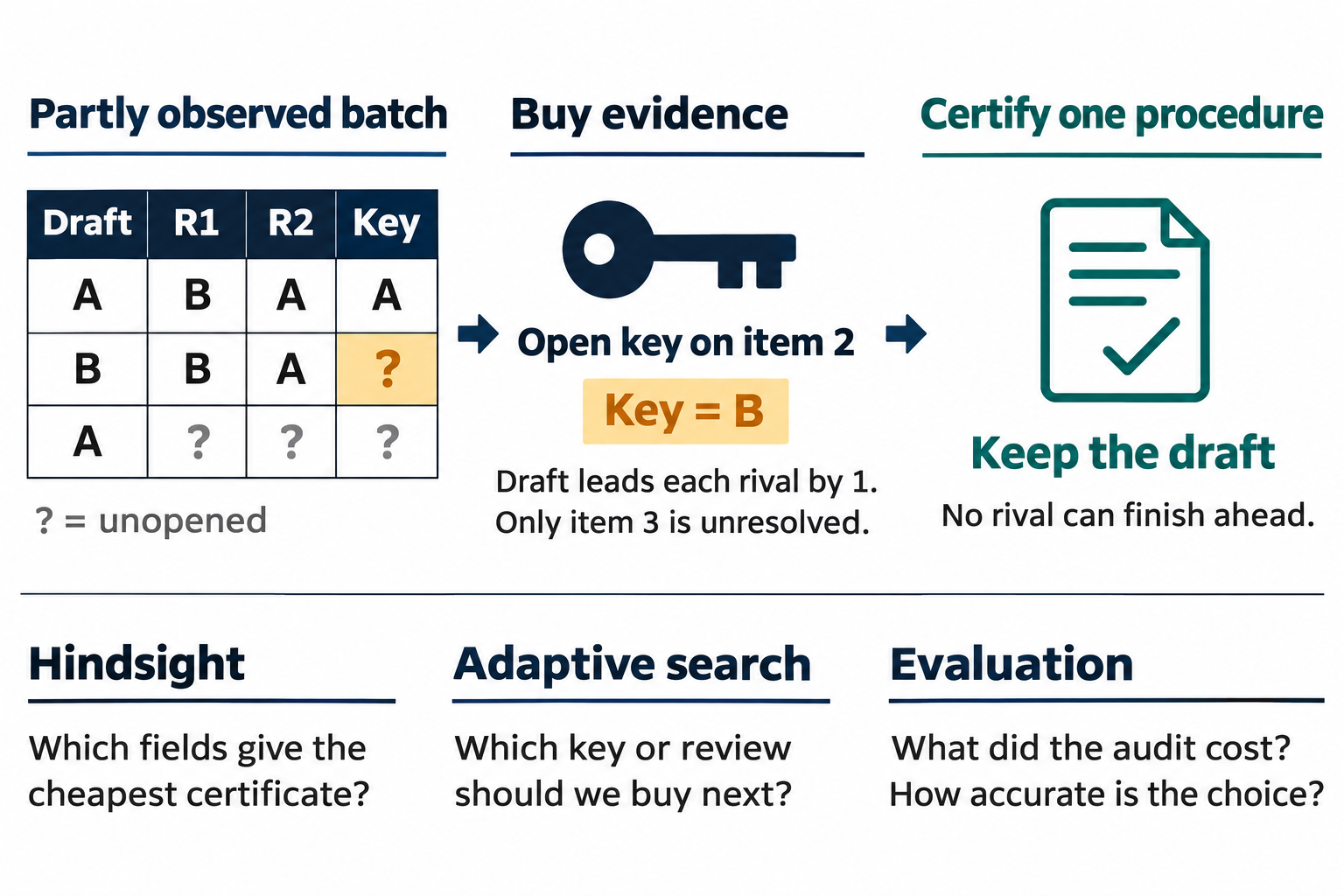}
\caption{Reviewer selection from partial evidence. Each row is an item. R1 and R2 are the answers two reviewers would release after reading the draft. The top row of panels shows one possible audit ending in an exact choice ($h=0$). The questions below connect this decision to our analysis of certificate cost and adaptive search, and to our evaluations on MMLU questions, numerical reports and Singapore electricity forecasts. Accuracy on new items is measured separately from the guarantee on the audited batch.}
\label{fig:auditoverview}
\end{figure}

Across a batch of reports the decision changes in one important way. The group must choose \emph{one procedure for the entire batch}, so a lead earned on one report can absorb a possible loss on another. Figure~\ref{fig:auditoverview} illustrates the stopping decision with three items. Opening the second key puts the draft one correct answer ahead of each reviewer. The last item can erase that lead, but neither reviewer can overtake the draft. The group can keep the draft on all three items without opening another field. We call the evidence justifying such a decision a \emph{certificate}. More generally, the group may allow a shortfall of $h$ correct answers. The audit can stop as soon as it can name a procedure that finishes within $h$ of the best on the batch for every possible value of the missing evidence. That distinction matters in practice. In our Singapore electricity study, rechecking just the five reports on which two procedures disagreed proved an exact tie across all 48. Recomputing all 48 showed that each procedure was correct on only eight. The first audit answered the group's question about which procedure to use. Discovering how often both procedures made the same mistake required checking their agreements too.

What changes when the group must pay for reviews as well as keys? Kuboyama \cite{kuboyama} studies a fixed batch in which every candidate's predictions are already recorded and the auditor chooses which labels to obtain. We begin with that case in Section~\ref{sec:keys}, where we give an exact stopping test and an integer program for the fewest keys that could certify a choice. With a price on each review, a disagreement itself may cost money to discover. We therefore distinguish two questions. An evaluator with the completed record can ask which pieces of evidence would demonstrate the choice as cheaply as possible. An auditor looking at a partly empty record must decide which field to buy next, before knowing what it contains. Both face the same standard of proof. How much extra can the search cost?

Our main theoretical results quantify this difference (Section~\ref{sec:paid}). Given the completed record, we group fields on the same item into bundles so that a key and the reviews it helps score can be valued together. For $K$ procedures with categorical answers, this gives a polynomial-time algorithm whose certificate costs at most $H_{2(K-1)}$ times the cheapest one, where $H_m=\sum_{r=1}^m 1/r$. The factor grows logarithmically with the number of procedures, independently of batch size and prices, and its leading logarithmic dependence is unavoidable unless P equals NP (Theorem~\ref{th:paidcover}). Searching for the evidence adds a different difficulty even with just two procedures. We construct a family with $n$ unresolved items where one purchase suffices in hindsight, yet the best valid search policy needs exactly $(n+1)/2$ purchases in worst-case expectation (Theorem~\ref{th:needle}). For two binary procedures, when items are independent and identically distributed under a known law, five counts summarize everything needed to minimize expected spending. The resulting policy uses the law to choose purchases, while its stopping guarantee holds for every completion (Theorem~\ref{th:bell}).

We evaluate the practical stakes by replaying saved outputs and revealing only the evidence available to the auditor at each decision (Sections~\ref{sec:design} and~\ref{sec:results}). With reviews already on file, the targeted audit used 222.7 keys on average per MMLU menu, compared with 239.2 for the best fixed order selected on development data. The chosen procedures averaged 2.94 percentage points higher held-out accuracy than choosing by stand-alone scores. On newly generated numerical reports the audit used 157.3 keys against 159.2 for the fixed order. An entropy-based baseline used fewer, and the held-out accuracy gain over stand-alone selection was zero. When both reviews and keys carried a price, the certificates constructed from completed records came within four price units of the exact minimum in all ten tested settings. These comparisons let us ask separately how much evidence the group needed to justify its choice and whether that choice improved the answers it released on new items.

\section{Setting and terminology}\label{sec:setting}
Picture a table with one row for each question and one column for each way the team could release an answer. An audit selects a single column for the whole batch. The team specifies how many correct answers it is willing to forgo relative to the best column, and the auditor must justify the choice using the evidence it has opened. Table~\ref{tab:terms} collects the notation for this decision.

Items are indexed by $t=1,\ldots,N$ and procedures by $i=0,\ldots,K-1$. Procedure $i$ releases the final answer $x_{ti}$ on item $t$. Procedure~$0$ releases the producer's original answer; every other procedure releases the answer of one reviewer after it has read the original. Until Section~\ref{sec:paid}, all final answers are assumed to be available, so only keys are bought. The key $y_t$ lies in a known finite set $\mathcal Y_t$. A common score $s_t(x,y)\in[0,1]$ defines what counts as correct. All experiments use exact match, and a malformed answer always scores zero. The score of procedure $i$ on the batch is $\bar s_i(y)=N^{-1}\sum_t s_t(x_{ti},y_t)$.

\begin{table}[htbp]
\centering\small
\caption{Terms used throughout the paper.}
\label{tab:terms}
\begin{tabular}{p{30mm}p{112mm}}\toprule
Term & Meaning \\ \midrule
Item & One question or report, indexed by $t$. A \emph{batch} is the fixed set of $N$ items on which a choice is made.\\
Producer, original & The analyst (a model) whose answer is checked, and that answer.\\
Procedure & A complete rule that releases one answer per item: keep the original, or release a named reviewer's answer after it has read the original. The \emph{menu} is the set of $K$ procedures.\\
Key & The verified correct answer for one item. One key scores every procedure on that item.\\
Field & Any purchasable piece of evidence on an item: its key or one procedure's answer (Section~\ref{sec:paid}).\\
Completion & An assignment of values to all fields not yet revealed, consistent with what has been revealed.\\
Tolerance $h$ & The total score (with exact match, the number of correct answers) the team is willing to forgo relative to the best procedure in the menu on this batch; $h=N\varepsilon$ for a tolerance $\varepsilon$ in score per item, rounded down when scores are whole answers.\\
Certificate & A set of revealed fields after which some procedure is guaranteed to be within $h$ of every other procedure for every completion. An \emph{exact} choice is one with $h=0$.\\
Valid policy & A rule for buying evidence that stops only when its current evidence is a certificate, whatever the completion.\\
Development, audit and held-out splits & In the experiments, items used to fit an audit rule, items on which keys are bought until a certificate is reached, and items used only to score the frozen choice.\\ \bottomrule
\end{tabular}
\end{table}

The benchmark for a choice is the best \emph{fixed} procedure in the same menu. A rule that sent each item to a different procedure would be a different decision and would need its own evaluation.

\paragraph{Why stand-alone accuracy is the wrong yardstick.}
A reviewer that reads the original can change its answer, so its stand-alone accuracy describes a different output from the one the procedure releases. In our experiments the released answer is the reviewer's read-first answer in full. Its number of correct answers therefore equals the original's, plus the original errors it repairs, minus the correct originals it spoils. Once the keys are known, these final answers determine the best procedure, and the pattern of shared errors adds nothing beyond them. If only some of a reviewer's changes are adopted, the same balance holds with adopted repairs and adopted spoils. The open question is how few keys suffice. Appendix~\ref{app:scope} gives the population form of this identity and treats procedures whose answers are themselves missing.

\paragraph{A worked example.}
Four items have final answers $(A,B,A,D)$ from reviewer~1 and $(A,C,B,D)$ from reviewer~2. Items~1 and~4 cannot change their relative score. Before any key is opened, either reviewer could lead by two. Suppose the key on item~2 is $B$. Reviewer~1 now leads by one on verified items, and on item~3 reviewer~2 can gain at most one, so reviewer~1 ties or wins whatever the remaining key. One key has settled the choice. Had the key on item~2 been $D$, both reviewers would have been wrong there and a second key would be needed. The value of the next key depends on the answer to the previous one.

\section{Certificates when only keys are bought}\label{sec:keys}
Suppose all reviewers have submitted their answers and only the answer keys cost money. The auditor can stop when no rival could gain enough from the unopened keys to overturn the choice by more than the allowed margin. We derive this stopping test, then use it to ask how few keys could have proved the same claim if their values had been known in advance.

Let $S$ be the set of items whose keys have been revealed. For a proposed choice $i$ and a rival $j$, define the rival's advantage on item $t$ under key $a$, its largest possible value, and the largest total lead the rival can still attain:
\begin{align}
 d_{tij}(a)&=s_t(x_{tj},a)-s_t(x_{ti},a), &
 b_{tij}&=\max_{a\in\mathcal Y_t}d_{tij}(a),\nonumber\\
 U_{ij}(S)&=\sum_{t\in S}d_{tij}(y_t)+\sum_{t\notin S}b_{tij}.
 \label{eq:bound}
\end{align}
The rival $j$ ranges over all procedures, including $i$ itself, for which $U_{ii}=0$. We assume that the unrevealed keys can take any combination of values. Known constraints linking keys across items could only tighten the bound.

\begin{theorem}[Exact certificate]\label{th:cert}
Given the available answers and the revealed keys, the largest shortfall of procedure $i$'s total score relative to the best procedure in the menu, over all completions of the unrevealed keys, equals $\max_j U_{ij}(S)$ (that is, $N^{-1}\max_j U_{ij}(S)$ in score per item). Some procedure is guaranteed to be within $h$ of the best for every completion if and only if $\min_i\max_j U_{ij}(S)\le h$. This holds after any sequence of keys, including one chosen adaptively from earlier keys.
\end{theorem}
\begin{proof}
Fix $i$ and $j$. Each unrevealed term is at most its maximum $b_{tij}$, and choosing a maximizing key on every unrevealed item attains the sum. Hence $U_{ij}(S)$ is the largest attainable lead of $j$ over $i$. Taking the maximum over the finitely many rivals and over completions in either order gives the largest shortfall of $i$, and taking the minimum over $i$ gives the stopping condition. The argument uses only the realized keys, so it holds for every adaptive sequence.
\end{proof}

In the worked example, reviewer~2's largest possible lead over reviewer~1 starts at two. Revealing key $B$ on item~2 turns that item's contribution from a possible $+1$ into a realized $-1$, which brings the largest remaining lead to zero. Revealing $D$ instead turns it into $0$ and leaves a possible lead of one. The two possible answers to the same query therefore lead to different stopping decisions.

The worst case for each rival may use a different completion, and these completions need not be attainable together. That does not weaken the test: a single rival under a single completion is enough to show that a proposed choice may exceed the tolerance. A failed test is therefore more informative than a wide error bar, because it exhibits a concrete set of key values under which the choice falls more than $h$ short of a rival.

\begin{theorem}[Minimum number of keys]\label{th:cover}
Fix the true keys $y$ and let $w_{tij}=b_{tij}-d_{tij}(y_t)$ be the reduction in rival $j$'s largest lead when the key on item $t$ is revealed. The minimum number $k^*(y)$ of keys that certify some procedure is
\[
 k^*(y)=\min_i\ \min_{z\in\{0,1\}^N}\left\{\sum_t z_t:\
 \sum_t w_{tij}z_t\ge\sum_t b_{tij}-h\quad\text{for every }j\right\},
\]
where an infeasible inner problem has value $+\infty$. Every valid audit policy opens at least $k^*(y)$ keys on this instance.
\end{theorem}
\begin{proof}
Write $z_t=\one\{t\in S\}$. Then $U_{ij}(S)=\sum_t b_{tij}-\sum_t w_{tij}z_t$, and Theorem~\ref{th:cert} gives one inequality per rival. Every feasible $z$ is a certificate of cost $\sum_t z_t$, and every policy that stops validly has opened a feasible set for the procedure it returns. The two minima are therefore equal, and no valid policy can stop sooner.
\end{proof}
Under exact match the reductions $w_{tij}$ are 0, 1 or~2. For a fixed choice $i$, items with identical reduction vectors can be grouped into one bounded integer variable, so $k^*(y)$ is cheap to compute for the menus studied here. The program uses the true keys, so it measures the cheapest certificate \emph{in hindsight}; it does not tell an auditor which keys to open.

\begin{proposition}[Searching costs more than proving]\label{prop:gap}
Consider two reviewers, binary answers and an exact choice ($h=0$). For every $m\ge1$ and every valid adaptive policy, there is an instance with $2m$ items and $k^*(y)=m$ on which the policy's expected number of keys is at least $2m-2m/(m+1)$. The ratio to the hindsight minimum therefore approaches two.
\end{proposition}
\begin{proof}
One reviewer always answers A and the other always B. Place $m$ keys of each kind in random positions; the policy must remain valid on every completion, including unbalanced ones. The reviewer answering A can be certified only once all $m$ keys equal to A have been seen, and symmetrically for B. A helpful observer could reveal $m$ identical keys at once, so $k^*(y)=m$. For any adaptive policy, the unopened positions are exchangeable under the random placement, so the sequence of opened keys up to the first possible stop is distributed as a prefix of a uniformly ordered balanced string. The earliest stop leaves exactly the final run of identical letters unseen. The $m$ A's fall exchangeably into the $m+1$ gaps between B's, so the expected number of trailing A's is $m/(m+1)$; trailing B's contribute the same. Subtracting the expected unseen run from $2m$ gives the bound. Since this is an average over instances, the policy's expected cost is at least this large on some instance; the argument also covers randomized policies.
\end{proof}
The gap to the hindsight minimum is therefore not only a weakness of a particular heuristic: it can remain for the best implementable search. The proposition is a lower bound on this overhead, not an approximation guarantee.

\subsection{Choosing the next key: the targeted audit}\label{sec:targeted}
When the stopping test fails, at least one rival could still finish far enough ahead to matter. The targeted audit gives priority to a key expected to narrow that rival's lead. Its estimate guides spending, while Theorem~\ref{th:cert} determines whether the evidence obtained is sufficient to stop.

At each round, let $i$ be the procedure minimizing $\max_j U_{ij}$, with ties broken by menu order. If $i$ is certified, return it. Otherwise compute each rival's outstanding excess $v_j=[U_{ij}-h]_+$. Let $p_t(a)$ be an estimated probability that the key on item $t$ is $a$. For each unopened item, compute
\begin{equation}
 q_t=\sum_j v_j\left[b_{tij}-\sum_{a\in\mathcal Y_t}p_t(a)d_{tij}(a)\right],
 \label{eq:query}
\end{equation}
the expected reduction in the rivals' leads, weighted by how far each rival still exceeds the tolerance. Open the key with the largest $q_t$, subtract the realized reduction from $U_{i'j'}$ for every pair of procedures $(i',j')$, and repeat. Ties among items are broken by a seeded random order. The implementation stores $O(K^2)$ pairwise bounds and $O(NK^2)$ per-item terms, and spends $O(NK)$ work per round after a one-time computation of expected reductions.

The current choice $i$ can change after any key. We estimate $p_t$ on a separate development sample. The answers on an item are encoded by the order in which letters first appear across the menu; for each resulting pattern we count how often each position holds the correct answer and add one to every count. Patterns not seen in development receive a uniform estimate, and letters predicted by no procedure are ordered by their original index. The estimate affects only the order of purchases; the validity of the returned choice follows from Equation~\eqref{eq:bound} alone. The score is a heuristic without a proven approximation ratio. In particular, it can undervalue keys that matter for a procedure that only later becomes the leading choice. An ablation replaces $p_t$ by a uniform distribution and so needs no development keys.

\paragraph{Using the audit on a new batch.}
For an exact-match task the auditor needs only the following steps.
\begin{enumerate}
\item Provide one entry per item for each procedure, with all answers in a common answer space, and choose the tolerance $h$.
\item Compute each rival's largest possible lead. If some procedure is within $h$ of every rival, return it together with the keys used and the remaining bound for each rival. This can happen before any key is opened, for example when all procedures agree everywhere.
\item Otherwise open the key chosen by Equation~\eqref{eq:query}, score every procedure on that item, update the bounds and return to step~2. If the budget runs out first, the current bounds state how large the unresolved shortfall can still be.
\end{enumerate}
For example, with a tolerance of seven answers on a 700-item batch, a bound of eight against even one rival leaves the choice uncertified, while a bound of at most seven against every rival certifies it. Both statements are counts on this batch and require no probability model of future items.

\section{Certificates when reviews and keys are both priced}\label{sec:paid}
A team that has only the original answers faces a choice at every purchase. It can commission a review to learn what a second procedure would release, or it can buy a key to score every answer already on file for that item. The value of one purchase can change after the other. This interaction matters both for constructing a cheap certificate in hindsight and for finding it during an audit.

Take an item on which the original says A and a reviewer would say B, under a binary answer set. Before anything is bought, the reviewer could lead by one on this item. Buying the review alone leaves that bound at one, because the key could still be B. Buying the key alone and finding A lowers it to zero. Buying both lowers it to minus one. The review is worth nothing on its own but worth one after the key, so its value \emph{increases} with other purchases. The standard greedy argument for covering problems assumes the opposite (diminishing returns), so it cannot be applied to individual fields.

\paragraph{Model.}
Let $L$ be the set of fields known at the start, on all items. On each item, the purchasable fields are its key and the missing answers of the procedures. Each field has a strictly positive price, prices add up, and each field is paid for once. A procedure's answer is fixed in advance, whether or not it is bought, and buying a field reveals its value. Given the revealed fields, a certificate must hold for every combination of the remaining values within the answer sets; no probabilistic or logical restrictions are imposed. In this section the score is exact match, every answer set has at least two elements, and the tolerance $h$ is an integer number of answers.

\paragraph{Evidence from a bundle of fields.}
Fix a completed record $\omega$, which a retrospective evaluator can inspect. For an item $t$, a proposed choice $i$, a rival $j$ and a set $B$ of that item's missing fields, let $\mathcal W_t(B)$ be the completions of the item that agree with $\omega$ on the initially known fields $L$ and on the fields in $B$. Define the rival's largest possible advantage on the item after buying $B$, and the reduction that $B$ achieves:
\[
 u_{tij}(B)=\max_{\mathcal W_t(B)}
  \big[\one\{x_{tj}=y_t\}-\one\{x_{ti}=y_t\}\big],
 \qquad w_{tij}(B)=u_{tij}(\varnothing)-u_{tij}(B).
\]
When all answers are available and $B$ is the key, $w_{tij}$ is the reduction of Theorem~\ref{th:cover}. By the same argument as in Theorem~\ref{th:cert}, $\sum_t u_{tij}$ is the exact largest lead of $j$ over $i$. Each $u_{tij}$ can be evaluated by trying every admissible key: once the key is fixed, an unknown rival answer can be correct and an unknown answer of $i$ can be wrong. Set $u_{tii}=0$. The reduction that procedure $i$ still needs against rival $j$, and its total over rivals, are
\[
 \Delta_{ij}=\Big[\textstyle\sum_t u_{tij}(\varnothing)-h\Big]_+,\qquad \Delta_i=\sum_{j\ne i}\Delta_{ij}.
\]
A choice of bundles $B_t$, one per item, certifies $i$ exactly when $\sum_t w_{tij}(B_t)\ge\Delta_{ij}$ for every rival $j$. Let $C^*$ be the smallest total price of a certificate for \emph{any} procedure that can be certified. A cheapest certificate need not buy both the review and the key on every item it touches.

\begin{theorem}[Approximating the cheapest certificate]\label{th:paidcover}
For $N$ items and $K\ge2$ procedures, a polynomial-time algorithm that inspects the completed record constructs a valid certificate of price at most $H_{2(K-1)}\,C^*$, where $H_m=\sum_{a=1}^m 1/a$. For a fixed procedure $i$ that can be certified, the factor is $H_{\min\{\Delta_i,2(K-1)\}}$ when $\Delta_i>0$; if $\Delta_i=0$, its certificate is free. Each greedy step examines at most four patterns of shared fields per item and sorts at most $K-1$ ratios of gain to price per pattern. For fixed $K$, an exact dynamic program runs in time polynomial in $N$. When $K$ is part of the input, approximating $C^*$ within a factor $(1-\delta)\ln(K-1)$, for any fixed $0<\delta<1$, is NP-hard, even with binary answers and every review already available.
\end{theorem}

The difficulty is deciding which fields to buy \emph{together}. Treating every subset of an item's missing fields as a single action restores diminishing returns, but there are up to $2^{K+1}$ such subsets per item. The proof in Appendix~\ref{app:costproof} handles both issues. It credits each rival with only the best reduction obtained on an item, so overlapping bundles cannot count the same evidence twice. It then uses the fact that, once we fix whether to buy the key and the proposed choice's answer (the two fields shared by every comparison involving that choice), each remaining reviewer answer affects only its own pairwise comparison. The best bundle can then be found by sorting. The approximation factor depends only on the number of procedures, not on the batch size or the prices, and its logarithmic order matches the hardness bound unless P equals NP. The theorem concerns the cost of a certificate constructed \emph{with hindsight}. The next result shows why that cost can be a misleading target for an auditor who must search.

\begin{theorem}[Finding evidence can cost linearly more]\label{th:needle}
For every $n\ge2$ there is a family of instances with two procedures, binary answers, unit prices, $n$ unresolved items and $h=0$, in which every instance has $C^*=1$. For every policy that is valid on all completions, including randomized policies, some instance in the family requires at least $(n+1)/2$ purchases in expectation. This is the exact minimax expected cost over the family.
\end{theorem}
\begin{proof}
Include $n-1$ fully revealed items on each of which the original is correct and the reviewer is wrong. On each of the $n$ unresolved items, the original says A and the hidden key is B. Every hidden reviewer answer is B except one, which is A. Buying that exceptional answer certifies the original: the exceptional item contributes zero, and the other $n-1$ unresolved items can offset at most the $n-1$ revealed losses. Before any purchase, either procedure can still lose, so the cheapest certificate costs exactly one.

Place the exception uniformly at random. Until it is found, every key bought is B and every review bought is B. A completion in which every remaining reviewer answer is B, and correct, still defeats the original. Certifying the reviewer instead requires at least $n-1$ items with both fields revealed, because only a revealed disagreement with a revealed key in the reviewer's favour has a strictly positive minimum contribution; this costs at least $2(n-1)\ge n$. Let $\kappa$ be the position of the exception in the policy's order of reviews on the all-B transcript, extended arbitrarily after the policy stops. If the policy finds the exception, it has spent at least $\kappa$; if it stops without finding it, it has spent at least $n\ge\kappa$. Given the policy's random seed, $\kappa$ is uniform, so the expected cost is at least $(n+1)/2$ on some instance. Conversely, reviewing items in uniformly random order until the first A certifies the original on every instance of the family at expected cost $(n+1)/2$. If all reviews are B, the policy falls back to opening keys until certified, which keeps it valid outside the family and proves the matching upper bound.
\end{proof}

\subsection{An optimal policy for two binary procedures}\label{sec:online}
An auditor in the search problem of Theorem~\ref{th:needle} knows little about which purchase is likely to settle the decision. If the team has an explicit model for disagreements and errors, it can use that model to minimize expected spending. For two procedures with binary answers, the remaining choices can be summarized by five counts, while the stopping decision still has to be justified for every completion.

Suppose the original answer on each item is visible and binary. Let $D=1$ if the hidden reviewer answer disagrees with the original, and $Z=1$ if the original is wrong. Buying the review reveals $D$; buying the key reveals $Z$. The reviewer's advantage on the item is $D(2Z-1)$. Assume that, given the visible originals, the pairs $(D,Z)$ on unresolved items are independent and share a known joint law that gives positive probability to all four values. The two bits of a pair may be strongly dependent. A key costs $c_Y$ and a review $c_R$ on every item.

For a single item with $h=0$ and no lead from other items, buying the key settles the choice immediately: keep a correct original, or choose the reviewer if the original is wrong. Buying the review first ends the audit at no further cost if it agrees; if it disagrees, the key is still needed. The exact expected price is therefore $\min\{c_Y,\;c_R+\Pr(D=1)c_Y\}$, and the better first purchase switches at $c_R=\Pr(D=0)c_Y$. With several items, a favourable outcome on one item can offset a loss on another, so applying this one-item rule to each item separately can waste information.

\begin{theorem}[Optimal cost from five counts]\label{th:bell}
For $n$ initially unresolved items and any initial lead, an optimal policy that is valid on all completions, under the specified law, can be computed in $O(n^5)$ arithmetic operations and space. Its state consists of five counts: the number $a$ of untouched items; the number $b$ of items with a known disagreement and an unknown key; the numbers $p$ and $m$ of items with a known key ($Z=1$ and $Z=0$ respectively) and an unknown review; and the reviewer's lead $\sigma$ on fully resolved items. The policy stops exactly when
\[
             \sigma-a-b-m\ge -h\quad\text{or}\quad \sigma+a+b+p\le h,
\]
choosing the reviewer in the first case and the original in the second. Otherwise at most five types of purchase need to be compared. A policy planned with a different full-support law keeps the same stopping guarantee; only its expected-cost optimality depends on the law used for planning.
\end{theorem}
\begin{proof}
Untouched items and items with a known disagreement each contribute a reviewer advantage in $[-1,1]$, items with $Z=1$ contribute in $[0,1]$, and items with $Z=0$ contribute in $[-1,0]$. All endpoints can be attained together, which gives the two stopping inequalities. Given the state, items of the same type have the same conditional law and independent remaining values, so the price plus expected continuation cost of every purchase depends only on the five counts. Table~\ref{tab:bell} lists every transition. Each strictly decreases $2a+b+p+m$, so backward induction gives the minimum over all adaptive policies, including those that use the full history. There are $O(n^4)$ count tuples and $O(n)$ reachable leads. The stopping inequalities use only the possible values of the fields, so they remain valid if the law is misspecified.
\end{proof}

\begin{table}[htbp]\centering\small
\caption{Transitions of the optimal policy. Apply the indicated changes to $(a,b,p,m,\sigma)$ and leave the other counts unchanged. For a purchase on a partly observed item, use the joint law conditioned on the bit already observed. A purchase type is available only when its count is positive.}\label{tab:bell}
\begin{tabular}{lll}\toprule
Purchase & Outcome 0 & Outcome 1 \\ \midrule
Review on untouched item ($D$) & $a-1$ & $a-1,\ b+1$ \\
Key on untouched item ($Z$) & $a-1,\ m+1$ & $a-1,\ p+1$ \\
Key after a disagreement ($Z$) & $b-1,\ \sigma-1$ & $b-1,\ \sigma+1$ \\
Review after $Z=1$ ($D$) & $p-1$ & $p-1,\ \sigma+1$ \\
Review after $Z=0$ ($D$) & $m-1$ & $m-1,\ \sigma-1$ \\ \bottomrule
\end{tabular}
\end{table}
The implementation evaluates $J(x)=\min_q\{c_q+\sum_o P(o\mid x,q)J(x_{q,o})\}$ outside the stopping set and sets $J=0$ inside it, where $q$ ranges over purchase types and $o$ over their outcomes. The known-law calculation in Section~\ref{sec:paidonline} uses the same law for planning and evaluation. The held-out replay estimates the law from development items, and the next purchase depends only on fields already bought.

\section{Experimental design}\label{sec:design}
The experimental design follows a review decision from calibration through evaluation. Studies~1 and~2 use separate items to fit the audit rule, choose a procedure, and measure its accuracy on new questions. Study~3 asks whether a choice that is justified on the audit batch also produces reports accurate enough to release.

Sections~\ref{sec:paidreplay} and~\ref{sec:paidonline} then hide reviews as well as keys in the same records, compare the mixed-certificate algorithm with exact optima, and test the binary policy on a separate threshold decision. These priced replays were designed after the analyses with available reviews.

\subsection{What the auditor sees}
For the studies with available reviews, the auditor starts with every procedure's answer on each item. Opening one key scores all procedures on that item. The evaluator holds the remaining audit keys separately and uses them after stopping to check the guarantee. This division lets us replay the choices a team could make before paying for more evidence.

\begin{table}[htbp]
\centering
\caption{How items are used. Counts are distinct items per split. The four MMLU menus use the same questions. The Singapore replay makes a single decision on one batch.}
\label{tab:design}
\begin{tabular}{lrrr}\toprule
Use of the item & MMLU & Numerical & Singapore \\ \midrule
Development: fit the audit rule and select the fixed order & 300 & 128 & --- \\
Audit: open keys until a procedure is certified & 700 & 256 & 48 \\
Held-out: score the frozen choice & 1,000 & 512 & --- \\ \bottomrule
\end{tabular}
\end{table}

Table~\ref{tab:design} shows how keys are used. On the development split, the audit rule may see all keys. On the audit split, every acquisition order starts from the same hidden keys and stops under the same certificate. The held-out keys are opened to the evaluator only after the choice is frozen, and we never replace a chosen procedure because another looks better on held-out items. A saving of 20 audit keys and a gain of two points in held-out accuracy are therefore different quantities.

\subsection{Study 1: reviewing another model's answer on MMLU}
Does a model that solves questions well on its own also make a good reviewer after reading another model's answer? The MMLU archive lets us compare those two roles on the same 2,000 multiple-choice questions \cite{mmlu}. An earlier study drew the questions from the subset that between 10\% and 85\% of a 79-model panel answered correctly. This sampling favours disagreement, so the reported accuracies describe this subset.

On every question, each of four models first answered independently and gave a short justification, and each of the other three then reviewed that answer. This gives 8,000 independent answers and 24,000 reviews. With Haiku-5-5 as the producer, for example, the menu has four procedures: keep Haiku-5-5's answer, or release the read-first answer of Haiku-4-5, GPT-5.4-mini or GPT-5.4-nano. Each of the four models serves once as producer, giving four menus. Model names follow the archive's identifiers. Answers are matched by producer, reviewer and question, and a malformed answer receives no credit. A fixed hash of the question identifier assigns each question to the same split in every menu, so a development question for one producer never appears in another producer's held-out split.

This analysis is retrospective: the investigators had examined aggregate results from these records before. In addition, each independent answer was generated in the producer role, whereas each reviewer saw both an answer and a justification. A difference between the two measures the difference between complete procedures and cannot isolate the effect of seeing the answer alone.

\subsection{Study 2: checking numerical reports with a local model}
A numerical report gives us an independent way to establish the right answer by repeating the calculation. We use that feature to study review under controlled conditions: a single local model checks short tables using three different reviewer instructions. This lets us inspect exactly which changes helped and which spoiled a correct figure.

We generated 896 tables, each with four rows of actual and forecast values. With differences $d_r=\text{actual}_r-\text{forecast}_r$, each item requests one of four statistics: the mean absolute difference $\sum_r |d_r|/4$, the largest absolute difference $\max_r |d_r|$, the signed sum $\sum_r d_r$, or the number of underforecasts $\sum_r\one\{d_r>0\}$. Each statistic makes up one quarter of every split. The four answer options are the correct value and the values of the other statistics, supplemented by nearby values when two statistics coincide. A deterministic calculation supplies each key independently of the model. Every table has its own seed, and all 896 tables are distinct.

Each table required seven calls to a local model. The first gives the producer's answer. Three reviewer configurations then answer twice: once on their own and once after seeing the value the producer chose. The configurations ask the reviewer, respectively, to perform the calculation, to check the definition of the statistic, or to work row by row while checking signs and units. All use the same Qwen2.5-Coder-1.5B-Instruct weights (Q4\_K\_M) \cite{qwen} at temperature zero, with output constrained to the letters A--D. Each reviewer sees the same option order in both of its conditions; orders may differ between reviewers. Letters are mapped back to numeric values before answers are compared. No justification is generated in this study. Because the three reviewers share one model, they are configurations rather than independent models.

\paragraph{One held-out item.}
On item 404, the actual values are $(26,31,54,12)$ and the forecasts $(30,35,50,16)$. The differences are $(-4,-4,4,-4)$, so the mean absolute difference is~4. The options offered are $5,4,1,-8$. The producer answered~5. After reading that answer, the calculation reviewer returned~4, the definition reviewer~5 and the row-by-row reviewer~1. One computation of the key therefore scores the four procedures $(0,1,0,0)$; different wrong answers receive the same score. This item was chosen after evaluation, for illustration.

The splits, generation protocol, primary comparator and one-point tolerance were recorded locally before any calls were made. Fourteen tables contain a forecast of zero or $-1$, because the signed offset can exceed a small actual value, whereas the design had specified positive inputs. The primary analysis keeps every table. A sensitivity analysis removes these tables from all splits, leaving 124, 250 and 508 items, and again finds no difference in held-out accuracy.

\subsection{Study 3: rechecking forecast reports on Singapore electricity data}\label{sec:study3}
Two reporting procedures can publish the same incorrect figure. To examine that possibility in a research workflow, we replay reports from an earlier pilot in which a local model analysed half-hourly demand in Singapore's National Electricity Market \cite{ema}. The public measurements allow us to compare the procedures on the rows where they disagree and to recompute the accuracy of each procedure over the entire batch.

The eight evaluation weeks start between 28 July and 22 September 2025. Each week has six reports, one for each four-hour block of the day; a block covers eight half-hour intervals on each of seven days. The correct value for a report is the mean absolute difference between forecast and actual market demand over those 56 records, rounded to 0.01~MW, which we recompute from the source measurements. We use the reports the model produced and exclude the faults that the earlier pilot injected on purpose.

In the pilot, the model chose a calculation plan, ran it in a deterministic notebook and then chose a figure to quote in its report. The two procedures compared here either keep the quoted figure or copy the result of the most recent notebook run. Both figures are visible before the source records are consulted. Copying the latest result can repair a stale quote, but the calculation itself can still use the wrong statistic. The study asks for an exact relative choice over all 48 reports under exact match. It has no development or held-out split.

\subsection{Audit orders, comparators and uncertainty}
We compare acquisition rules by charging each for the keys it opens to reach the same tolerance. We then freeze the selected procedure and score it on separate questions. A rule can save keys and still select the same reviewer as a more expensive rule, so key cost and subsequent accuracy are reported separately.

The primary comparison is against the best of three simple acquisition orders, chosen on development data. \emph{Uniform} opens any unopened key at random. \emph{Uniform on disagreements} first opens items on which at least two answers differ. \emph{Vote disagreement} gives priority to items that separate more pairs of procedures. We freeze the order that uses the fewest development keys on average over 20 seeds. Vote disagreement is chosen for all four MMLU menus and uniform on disagreements for the numerical reports. Every audit allows a shortfall of one percentage point from the best procedure, that is, at most seven correct answers on 700 items or two on 256.

The other orders isolate components of the method. \emph{Static expected reduction} sums the expected reductions over all pairs once, without updating the current choice. The \emph{uniform-prior} ablation keeps the targeted rule but learns no probabilities. \emph{Model Selector} opens the key that minimizes the expected entropy over which model is best \cite{selector,coda}; Appendix~\ref{app:replay} gives our implementation. For comparison with the closest fixed-pool work, we adapt Kuboyama's surviving-pair width rule \cite{kuboyama} to categorical answers: a procedure leaves the query score once a rival beats it under every completion, and each unopened item is scored by the sum of $b_{tij}+b_{tji}$ over pairs of surviving procedures. All procedures remain in the stopping test. This comparison uses the same answers and tie seeds as the primary one and was added after the original study.

To score the chosen procedure on held-out items, we compare five ways of choosing. \emph{Stand-alone} ranks reviewers by their independent accuracy on the audit split with every key opened and then deploys the chosen reviewer in the read-first condition (or keeps the producer if it ranks first). \emph{Targeted} is the procedure certified by the targeted audit. \emph{Full} ranks procedures by their final answers on the audit split with every key opened. \emph{Development-only} does the same on the development split alone. \emph{Wider} is Full with a larger menu that also allows deploying a reviewer's independent answer. Finally, \emph{Oracle} reports the best procedure in the original menu as judged on the held-out items themselves; it shows the available headroom and is not a usable rule.

Each acquisition order is run with 20 seeds for breaking ties, paired across orders. We average over seeds and then weight each producer equally. For held-out uncertainty, we compute the paired difference in correctness between two ways of choosing on each question, average it over seeds and menus, and resample whole questions 4,000 times. The percentile interval keeps the dependence between producers that share a question. It reflects variation over held-out questions, conditional on the models, the fitted development rule and the audit split; the tie seeds are repeated runs on the same data.

\section{Results}\label{sec:results}
Across the two main datasets, disagreements let an auditor justify a choice while leaving some keys unopened. Whether that choice helps on new items depends on the task. We report the price of the decision and its held-out accuracy, use the Singapore reports to examine shared errors, and then consider what happens when the reviews themselves also carry a price.

\subsection{Keys needed for a one-point certificate}
How much does query order matter when every policy must justify the same choice tolerance? Table~\ref{tab:cost} counts the keys each policy opens before it can certify a procedure within one percentage point of the best on the audit batch. The hindsight minimum shows how much evidence would have been sufficient if the auditor had known where to look.

\begin{table}[htbp]
\centering
\caption{Mean number of keys opened until a one-percentage-point certificate (lower is better). Development keys are excluded. The last row is computed with all keys known and is an offline lower bound, not a usable order. Differences quoted in the text are computed before rounding.}
\label{tab:cost}
\begin{tabular}{lrr}\toprule
Acquisition order & MMLU ($N=700$) & Numerical ($N=256$) \\ \midrule
\tableinput results/audit_table.tex
\bottomrule\end{tabular}
\end{table}

\begin{figure}[htbp]
\centering\includegraphics[width=.93\linewidth]{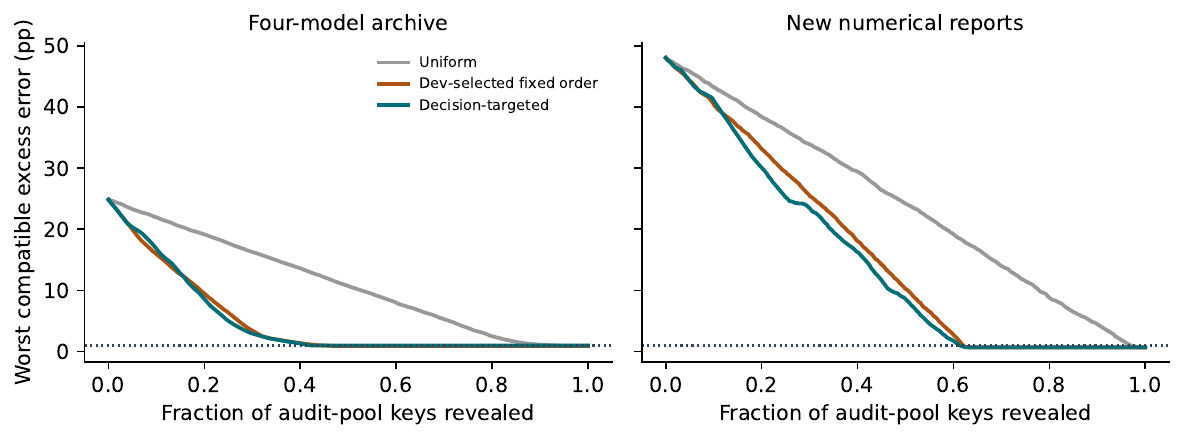}
\caption{Largest possible shortfall of the current choice as keys are opened, averaged over menus and tie orders. In the figure's labels, ``worst compatible excess error'' is this shortfall, ``Decision-targeted'' is the targeted audit, ``Dev-selected fixed order'' is the development-selected fixed order, ``audit-pool keys'' are keys of the audit split, and ``Four-model archive'' and ``New numerical reports'' are Studies~1 and~2. Each curve stays at its final value after it stops. The dotted line marks one percentage point. The fixed order is chosen separately on each development split.}
\label{fig:curve}
\end{figure}

Table~\ref{tab:cost} gives the cost of reaching a certificate, whichever procedure wins. On MMLU, the targeted audit uses 16.5 fewer keys than the development-selected fixed order (6.9\%) and 27.9 fewer than Model Selector. The cheapest certificate in hindsight costs 145.25 keys on average, so the targeted audit spends 1.53 times this lower bound. On the numerical reports the saving over the development-selected order is 1.85 keys (1.2\%), and Model Selector uses 2.95 fewer keys than the targeted audit. Every order that skips agreements beats uniform sampling by a wide margin, so most of the advantage over uniform sampling comes from the answers alone. Figure~\ref{fig:curve} shows how quickly the remaining shortfall closes.

\paragraph{Where the savings come from.}
The four MMLU menus have between 351 and 433 unanimous items out of 700 (Table~\ref{tab:width}). No key is needed on these items to compare procedures. The numerical audit split has only 92 unanimous items out of 256, so most items give some pair a chance to gain. Among the items with disagreements, what matters is whether they separate procedures that can still win. The surviving-pair order tracks this as the audit proceeds. On MMLU it uses 231.85 keys per menu on average, against 222.69 for the targeted audit; on the numerical reports it uses slightly fewer, 156.55 against 157.30. The targeted audit is cheaper on all four MMLU menus, but only slightly for the GPT-5.4-mini and GPT-5.4-nano menus. Comparing only with uniform sampling would hide how much of the saving the answers themselves already provide.

\begin{table}[htbp]
\centering
\caption{Which disagreements matter. Rows are menus, named by producer. Unanimous items need no key for a relative choice. The third column gives the mean number of keys for the surviving-pair order; the last gives the range of key counts for the targeted audit over 20 tie orders on the same audit split.}
\label{tab:width}
\begin{tabular}{lrrr}\toprule
Menu & Unanimous items & Surviving-pair keys & Targeted range \\ \midrule
\tableinput results/survivor_table.tex
\bottomrule\end{tabular}
\end{table}

\paragraph{Is the learned estimate worth its development keys?}
Replacing the learned estimate with a uniform one costs only 2.1 extra keys per MMLU menu and 0.15 on the numerical reports. If no development keys are already available, the uniform version therefore costs 224.8 keys per MMLU decision, against $300+222.7$ for the learned version; the numerical totals are 157.45 against $128+157.3$. The comparison changes if development keys have already been paid for or are reused across decisions, but these data give little reason to buy a development set only to train the estimate. Relaxing the tolerance to five points reduces the targeted audit's cost to 174.0 keys on MMLU and 139.1 on the numerical reports; requiring an exact choice raises it to 235.2 and 160.3. On the numerical reports Model Selector is cheaper at tolerances of zero and one point and more expensive at five points (142.0). The release includes all three tolerances and their offline minima.

\subsection{Accuracy of the chosen procedure on held-out items}
A certificate settles a comparison on the audited questions. To see how well that decision travels, we keep the chosen procedure fixed and score it on questions that took no part in the audit. Each row of Table~\ref{tab:transfer} follows one producer through this comparison.

For each menu, the Stand-alone and Targeted columns choose from the same read-first menu in two ways. The Full column shows what the team would obtain by opening every audit key and ranking final answers directly. The Oracle column uses the held-out keys in hindsight. A choice can match Full and still fall short of Oracle, because the best procedure on one finite set of items need not be best on another.

\begin{table}[htbp]
\centering
\caption{Held-out accuracy (\%) of the chosen procedure, and audit costs, by menu (MMLU menus are named by producer). Stand-alone ranks reviewers by independent accuracy and deploys them read-first. Targeted uses the keys opened by the targeted audit. Full ranks final answers using all audit keys. Wider adds reviewers' independent answers to Full's menu. Oracle chooses with hindsight on the held-out items, within the original menu. $k^*$ is the hindsight minimum number of keys on the audit split, and Keys is the targeted audit's mean.}
\label{tab:transfer}
\small\begin{tabular}{lrrrrrrr}\toprule
Menu & Stand-alone & Targeted & Full & Wider & Oracle & $k^*$ & Keys \\ \midrule
\tableinput results/transfer_table.tex
\bottomrule\end{tabular}
\end{table}

\paragraph{MMLU.}
The targeted audit's choice reaches 79.69\% held-out accuracy, against 76.75\% for stand-alone ranking, a difference of 2.94 points (paired 95\% bootstrap interval 1.51 to 4.37). With Haiku-4-5 as producer, both rules keep the original. With GPT-5.4-mini as producer, the audit also keeps the original, which is 4.8 points more accurate than the reviewer chosen by stand-alone accuracy. With GPT-5.4-nano as producer, 11 of 20 tie orders choose Haiku-5-5 as reviewer and nine choose GPT-5.4-mini. Both satisfy the tolerance on the audit split, but their held-out accuracies differ by 2.5 points. Full averages 79.98\%, Development-only 79.40\% and Wider 80.30\%. The evidence supports judging a procedure by its final answers under the conditions in which it will run. It does not show that the adaptive order itself improves accuracy over other ways of using the final answers.

\paragraph{Numerical reports.}
On the numerical reports, the targeted audit and stand-alone ranking both reach 32.81\%, a difference of 0.00 points (paired interval $-4.64$ to $+4.45$). Of the 512 held-out tables, the original is correct on 154. The calculation and definition reviewers each raise this to 168: one repairs 79 answers and spoils 65, the other repairs 49 and spoils 35. The best procedure in hindsight, the row-by-row reviewer, reaches 175 correct (34.18\%), and choosing on the development split alone happens to select it. With four options and low absolute accuracy, exact numerical reporting is hard for this model, and there is little room to choose among its configurations. The interval allows effects in either direction, and these results provide no evidence of a held-out improvement.

\begin{table}[htbp]
\centering
\caption{Correct answers out of 128 held-out numerical reports per statistic. Each reviewer reads the original answer. Each statistic carries equal weight. The columns describe individual procedures; a choice still applies one procedure to the whole batch.}
\label{tab:operations}
\begin{tabular}{lrrrr}\toprule
Requested statistic & Original & Calculation & Definition & Row-by-row \\ \midrule
\tableinput results/operations_table.tex
\midrule Total correct (of 512) & 154 & 168 & 168 & 175 \\ \bottomrule
\end{tabular}
\end{table}

Table~\ref{tab:operations} shows how equal totals can hide different behaviour. The definition reviewer gets six more signed sums right than the calculation reviewer (46 against 40), while the calculation reviewer gets twelve more underforecast counts right (52 against 40). They finish level at 168. These are descriptive breakdowns of held-out outcomes. A rule that sent each statistic to a different reviewer would need its own calibration and evaluation.

\subsection{Study 3: a certified tie between two unreliable procedures}
For the electricity reports, a team could publish either the figure quoted in the report or the latest notebook output. We can settle which rule is more accurate by recomputing the cases on which the figures differ. Recomputing the cases on which the figures agree answers a second question that matters before release: how often did both rules repeat the same mistake?

\paragraph{One disagreement, from records to score.}
In the third block of the week starting 11 August, the report quoted 35.77~MW and the notebook printed $-99.36$~MW. Replaying the 56 measurements gives a sum of absolute forecast errors of 2,259~MW; dividing by 56 and rounding gives 40.34~MW. Both procedures score zero on this report. Rechecking it settles their relative score even though neither produced the correct figure.

\paragraph{What five checks establish.}
The two figures agree on 43 reports. On each of the other five, recomputation shows that both figures are wrong. These five checks (280 source records) certify a tie whatever the correct values of the other 43 reports. They are also necessary under our information model, which allows an unopened key to equal either submitted figure or neither. A team that also enforced domain constraints, such as nonnegative absolute errors, could sometimes rule out a figure without reopening the records. To assess absolute quality we recomputed all 48 correct values from 2,688 records. Of the 43 agreements, eight are correct and 35 are shared errors, so each procedure releases eight correct reports. The relative choice is now settled exactly; the reports themselves are mostly wrong, so a team deciding whether to release them should check the calculation upstream.

\subsection{Cheapest certificates when reviews are priced}\label{sec:paidreplay}
When both kinds of evidence have a price, the completed record lets us measure how close the bundle algorithm in Theorem~\ref{th:paidcover} comes to the least expensive valid certificate. We compare its purchases with an exact integer-program solution on the same saved answers. Both algorithms see the completed record, so this experiment measures the cost of exhibiting evidence in hindsight.

For each menu we take the first 64 items of the audit split in its fixed hash order. The producer's answer is visible; the three reviewers' answers and the key are hidden, and an exact choice is required ($h=0$). A key costs one and a review costs either one or four. These are illustrative prices: hiding a previously generated answer emulates the need to buy it. For each procedure that can be certified on the completed record, we solve the grouped integer program exactly and run the greedy algorithm of Theorem~\ref{th:paidcover}, and we report the cheapest certificate each method finds. For every greedy certificate, the release lists the fields bought and every rival's remaining largest lead.

\begin{table}[htbp]\centering
\caption{Price of a certificate for an exact choice on 64 items per menu (MMLU menus are named by producer). Revealing every field costs 256 when reviews cost one and 832 when they cost four. Minimum is the exact integer-programming optimum (the solver checks integer feasibility and agreement with its dual bound); Greedy is the algorithm of Theorem~\ref{th:paidcover}.}\label{tab:paidwitness}
\begin{tabular}{lrrrr}\toprule
& \multicolumn{2}{c}{Review price 1} & \multicolumn{2}{c}{Review price 4}\\
Menu & Minimum & Greedy & Minimum & Greedy\\ \midrule
\tableinput results/paid_hindsight.tex
\bottomrule\end{tabular}
\end{table}
In all ten settings (five menus at two prices), the greedy certificate costs at most four price units more than the exact minimum (Table~\ref{tab:paidwitness}). For GPT-5.4-mini it attains the minimum at both prices. The GPT-5.4-nano menu needs substantially more evidence on the same questions. These values show what a completed record can prove; they do not show what an auditor would have spent to find that evidence. Restricting the replay to 64 items makes an exact check possible but limits extrapolation to larger batches.

\subsection{Does the freedom to buy either field help?}\label{sec:paidonline}
An auditor making purchases in real time can choose the order of the two fields on each item. A cheap review might settle one item immediately, while a reference key might be the better first purchase on another. We measure what that flexibility buys when a policy must decide before it sees the missing values.

We compare the optimal binary policy of Theorem~\ref{th:bell} with two restricted versions solved by the same recursion. \emph{Key-first} must buy an item's key before its review; \emph{review-first} imposes the reverse order. Both may move between items and may stop without buying the second field when the evidence allows. A single fixed order over all fields would be a weaker comparator. All three policies require an exact choice ($h=0$), start without a lead, and are optimized within their own sets of actions at each price.

\begin{table}[htbp]\centering\small
\caption{Expected price per decision on eight items. Left: exact expected prices under a specified joint law. Right: mean realized prices over 16 held-out numerical batches, using a law estimated on development items. Every completed decision in the replay is checked against the full scores. ``Either'' may buy either field next.}\label{tab:paidonline}
\begin{tabular}{rrrrrrr}\toprule
&\multicolumn{3}{c}{Known-law calculation}&\multicolumn{3}{c}{Held-out replay}\\
Review price & Either & Key-first & Review-first & Either & Key-first & Review-first\\ \midrule
\tableinput results/paid_online.tex
\bottomrule\end{tabular}
\end{table}
For the known-law calculation we set the probabilities of $(D,Z)=(0,0),(0,1),(1,0),(1,1)$ to $(0.45,0.15,0.10,0.30)$ and $c_Y=1$, so that a disagreement signals a likely error in the original. When reviews cost four, for example, the policy free to choose either field has expected cost 25.302, against 26.533 for the better restricted policy (Table~\ref{tab:paidonline}). These values sum over all transitions of the recursion and carry no Monte Carlo error.

The held-out replay poses a different, genuinely binary decision on the numerical records: does the reported mean absolute difference exceed four units? We apply this strict threshold to the values chosen by the original and by the calculation reviewer, and to the independently computed key. This rescoring is exploratory and concerns a coarser decision than reporting the exact figure. We use only items requesting the mean absolute difference. On its 32 development items, the counts of $(D,Z)=(0,0),(0,1),(1,0),(1,1)$ are $(15,6,8,3)$, and we add one to each count before planning. The 128 corresponding held-out items form 16 consecutive batches of eight. Within each item type the code uses a fixed tie order, and all three policies receive the same records and prices. When reviews cost four, the policy free to choose either field, planned with the estimated law, costs 22.312 per batch, against 22.500 for key-first and 29.062 for review-first. When reviews cost 0.25, it matches review-first. Its gain over the \emph{better} restricted policy is modest here. Every decision remained valid despite the difference between development and held-out frequencies, because each stopping decision used the completion bounds rather than the estimated probabilities.

\paragraph{Checks of the theory code.}
Separate tests enumerate all compatible completed records for 36 small binary instances and 12 categorical instances with malformed answers, compare the minimum costs with exhaustive search over purchase sets in the binary cases, and compare the sorting step with explicit enumeration of bundles along the greedy paths actually taken. Twelve set-cover instances check the cost-preserving reduction. For the optimal policy, 24 comparisons with decision trees that use the full history check the five-count recursion under different prices, laws, tolerances and purchase restrictions. A separate exact search confirms the expected cost of the construction in Theorem~\ref{th:needle} for $n=2,3,4,5$. These checks test the implementation and small cases; the general claims rest on the proofs.

\section{Related work}
Model selection, priced search and active evaluation each address a part of the reviewer's decision. Work on fixed predictions provides the starting point for certifying a ranking. Work on priced queries helps explain why a short proof can be expensive to find. In our setting the auditor can buy a reviewer response and its key separately, and the value of one purchase depends on what the other reveals.

\paragraph{Choosing among models on a fixed pool.}
The closest formulation is Kuboyama's analysis of selective prediction \cite{kuboyama}, which certifies rankings for weighted binary errors on a fixed pool, including comparisons that share a multiclass prediction on each item. Section~4.5 of that work identifies differing multiclass predictions as a case that needs categorical labels. Numerical item 404 shows why: with reports of 5, 4 and~1, a single bit saying whether the producer is wrong cannot score the two alternatives. Maximizing the rival's advantage over the possible numeric keys gives the exact contribution of an item. Our key-only cover and the separation between search and hindsight follow the same principles as that analysis; Proposition~\ref{prop:gap} illustrates the search cost directly for reviewers.

\paragraph{Priced information and covering.}
The mixed-field result addresses a different source of difficulty. The classical greedy approximation for submodular cover is due to Wolsey \cite{wolsey}. In our problem, single field purchases can have increasing returns; the proof builds an equivalent cover over bundles of fields on one item and evaluates its greedy step without listing the bundles. Charikar et al.\ \cite{priced} compare the cost of adaptive queries with the cheapest proof on a realized input. Theorem~\ref{th:needle} embeds unstructured search in a two-procedure comparison and gives the exact value for its family; that search can cost more than proof is classical. Stochastic Boolean function evaluation also studies certificates and test ordering with priced observations \cite{sbfe}. The binary model of Section~\ref{sec:online} differs in allowing dependence between the two tests on an item, accepting either procedure in a tie, and pooling score differences across items. Reyzin et al.\ \cite{costexpert} study online expert advice with paid observations and cumulative loss, whereas our objective is the price of a single final certificate on a fixed batch.

\paragraph{Label-efficient model selection.}
Active model selection already addresses expensive labels for pretrained classifiers. Karimi et al.\ study sequential acquisition in a stream \cite{stream}. Model Selector chooses pool items to reduce uncertainty about which model is best \cite{selector}. CODA models dependencies among classifiers and classes \cite{coda}, and SELECT-LLM studies annotation-efficient selection of language models \cite{llmselect}. These methods make selection with few labels a demanding baseline. Our question is narrower: what can an auditor guarantee about a fully specified batch for every possible value of the unopened keys? An information-gain order can be combined with the same certificate, as our experiments do with Model Selector; CODA and SELECT-LLM provide broader context.

\paragraph{Teaching, partial identification and reliance.}
The cheapest certificate is related to the teaching dimension, in which a helpful teacher chooses examples that identify a target concept \cite{teach}. Here the keys can be inconsistent with every procedure, and a certificate need only establish a near-best score while leaving many possible truths. Decision-making under partial identification studies actions compatible with incomplete evidence \cite{cui}; our certificates apply this view to observable final answers and a purchasable unit of evidence. Research on reliance and team accuracy explains why helpful and harmful changes should be recorded separately \cite{reliance}; the repair and spoil counts here show which changes alter the final choice.

\section{Scope and implications}
A lab using an audit result needs to know which decision the evidence supports. A certificate applies to specified procedures on a fixed batch. Using a procedure on new inputs, estimating the cost of a search before it starts, and approving a report for release each require further assumptions or evidence.

The bounds assume that keys are verified correctly and that each procedure's answers are fixed. Mixed certificates further assume additive prices and that revealing one field does not change the value of another. If a query changes what a reviewer sees, its answer is the output of a different protocol and must be modelled as such. The polynomial-time optimal policy applies only to two procedures with binary answers and an exchangeable law with independent items; with heterogeneous reviewers or linked errors across steps of a workflow, its optimality argument does not apply.

In the experiments with available reviews, the held-out gain over stand-alone ranking depends on deploying reviewers in the read-first condition and on keeping the original in the menu. Allowing reviewers' independent answers to be deployed can improve the simple comparator. Regenerating stochastic answers, changing prompts or moving a reviewer to a new setting changes the decision being certified. The subject mix of the MMLU sample and the shared templates of the numerical study limit how far the results generalize. The bootstrap captures variation across items within each study; it does not capture uncertainty from the choice of models, model versions, development samples or templates.

For a team validating reports, the difference between a hindsight benchmark and an implementable policy matters. The greedy algorithm prices the evidence that a completed experiment contains. A team that must decide before that evidence exists can use the binary policy when its model fits, or the targeted audit when all reviews are already available. A policy with a comparable cost guarantee for categorical answers, several reviewers and online decisions under a stated population model remains open. Certifying that a procedure meets an absolute quality threshold is a different question, as the Singapore replay shows.

\section{Conclusion}
A team budgeting for review faces two costs. One is the price of the smallest body of evidence that would justify its choice on a completed batch. The other is the expense of locating enough evidence while the answers are still hidden.

The first can be approximated within a factor logarithmic in the number of procedures, for any prices and any batch size, and this factor matches the hardness bound to leading order. The second can grow linearly while the cheapest certificate stays fixed. For two binary procedures under an exchangeable law, the next purchase can be chosen optimally under a declared model while the final choice is still checked without trusting that model. The experiments show both the room for this distinction and its limits: constructed certificates come close to the hindsight minimum, while the freedom to choose which field to buy next gives only modest savings over the better of the key-first and review-first policies on held-out numerical batches.

\section*{Reproducibility}
The accompanying package lets a reader replay the audit in the order the evidence was opened and check the final choice against the complete record. It contains the archived answer letters, new prompts and outputs, local design record, split logic, deterministic replay code, integer solver and exhaustive checks on small instances.

No human participants were recruited. All new model calls used a local model. Key counts and stated review prices measure the information used in a replay; monetary savings and human verification time were not measured. The package also contains separate checks of the certificate optimization and of the online decision trees.

OpenAI Codex assisted with manuscript drafting and revision. Figure~\ref{fig:auditoverview} was created with OpenAI image generation.

\appendix
\section{Proof of Theorem~\ref{th:paidcover}}\label{app:costproof}
The main obstacle in the approximation proof is that buying two fields together can be more valuable than buying either alone. We first treat fields on the same item as a bundle and show how to credit overlapping bundles without counting their evidence twice. We then use the structure of a pairwise comparison to find the best bundle without enumerating every subset, and derive the hardness bound using set cover.

Fix a procedure $i$ that can be certified on the completed record. If no key or answer needs to be bought, there is nothing to prove. Define a virtual action $e=(t,B)$ for every nonempty bundle $B$ of missing fields on item $t$, with cost $c(e)$ equal to the total price of the fields in $B$. A set $A$ of virtual actions may contain several overlapping bundles on the same item. Its pessimistic evidence credit and utility are
\[
 M_{tj}(A)=\max\big(\{w_{tij}(B):(t,B)\in A\}\cup\{0\}\big),\qquad
 F(A)=\sum_{j\ne i}\min\left\{\Delta_{ij},\sum_t M_{tj}(A)\right\}.
\]
Revealing more fields can only narrow the possible completions, so $w$ is nonnegative and monotone in its bundle. Buying the union of the selected bundles on an item gives every rival at least $M_{tj}$. Thus $F(A)=\Delta_i$ implies a real certificate of no greater cost. Conversely, representing each item of a cheapest real certificate by a single bundle gives $F=\Delta_i$ at the same cost. The two optimal values are therefore equal, even though other virtual solutions may pay twice for a field or fail to credit a complementarity.

Each $M_{tj}$ is a maximum of nonnegative constants indexed by the selected actions; its marginal increment is the excess over the current maximum, truncated at zero. It has diminishing returns. Sums preserve this property, as does capping a nondecreasing utility by a fixed deficit: an increment is the lesser of its previous marginal and the remaining deficit, both nonincreasing as the selected set grows. Hence $F$ is a monotone integer-valued submodular cover objective. A bound in terms of the full goal $\Delta_i$ would grow with the batch size. We use the sharper integer submodular-cover charging argument of Wolsey \cite{wolsey}. Fix a cheapest virtual cover $o_1,\ldots,o_r$ for $i$; by the previous paragraph its total cost $C_i$ equals the price of the cheapest certificate for $i$. Let $A_\tau$ be the virtual greedy set just before step $\tau$, and $O_\ell$ the first $\ell$ optimal actions, with $O_0=\varnothing$. Allocate the residual among the optimal actions by
\[
 \beta_{\tau\ell}=F(A_\tau\cup O_\ell)-F(A_\tau\cup O_{\ell-1}).
\]
These nonnegative integer credits sum to $\Delta_i-F(A_\tau)$. Each decreases along the greedy trajectory by diminishing returns. If $\beta_{\tau\ell}>0$, action $o_\ell$ has current marginal at least $\beta_{\tau\ell}$, so the greedy price per unit gain is at most $c(o_\ell)/\beta_{\tau\ell}$. The gain at step $\tau$ equals $\sum_\ell(\beta_{\tau\ell}-\beta_{\tau+1,\ell})$. Charge each such decrement at the greedy unit price. The charge to action $o_\ell$ over all steps is at most $c(o_\ell)H_{\beta_{0\ell}}$, interpreting $H_0=0$: an integer drop from $\beta$ to $\beta'$ costs at most $c(o_\ell)(\beta-\beta')/\beta\le c(o_\ell)\sum_{q=\beta'+1}^{\beta}1/q$. Finally $\beta_{0\ell}\le F(\{o_\ell\})\le \min\{\Delta_i,2(K-1)\}$. The last bound is specific to exact match: even buying every field on an item reduces each rival's largest lead by at most two answers. Summing over the optimal actions proves the stated harmonic bound, independent of the number of items and of the prices. It remains to implement the greedy step, which ranges over exponentially many bundles.

Fix an item and the current credits. Deciding whether the bundle includes the key and whether it includes the answer of procedure $i$ gives at most four possibilities for this shared part $P$. Let $\rho_j$ be the residual requirement of rival $j$ after the current credits on all items. For a bundle $B$ on item $t$, its incremental utility is exactly
\[
 \sum_{j\ne i}\min\{\rho_j,[w_{tij}(B)-M_{tj}(A)]_+\}.
\]
With $P$ fixed, the summand for $j$ depends on only one further field, the answer of procedure $j$; the other answers do not enter this pairwise maximum. Write $g_0$ for the gain from $P$ alone and $g_j\ge0$ for the additional gain from including $j$. The gain per unit price of a subset $T$ of the rivals' missing answers is
\[
                 \frac{g_0+\sum_{j\in T}g_j}{c(P)+\sum_{j\in T}c_j}.
\]
At an optimal positive ratio $\lambda$, every available field with $g_j/c_j>\lambda$ must be included and every field with a smaller ratio can be dropped; both statements follow by cross-multiplication. Among fields with equal ratios a prefix can be chosen, keeping the bundle nonempty when the shared part is empty. Sorting the ratios and checking each prefix, together with $P$ when it is nonempty, therefore solves the greedy step on each item exactly. At most $\Delta_i\le N(K-1)$ greedy steps are needed, since each step raises the integer $F$ by at least one. Run the procedure for every procedure $i$ that can be certified and keep the cheapest resulting certificate. Applying the bound for fixed $i$ to a procedure with the globally cheapest certificate proves the approximation claim. Prices may differ across items and fields.

For the exact algorithm with fixed $K$, process the items in order and record, for each vector of reductions achieved against the $K-1$ rivals (component $j$ capped at $\Delta_{ij}\le N$), the minimum price paid so far. For each item, enumerate its at most $2^{K+1}$ bundles and add their reductions to update the table. After precomputing the reductions of each bundle, this takes $O(N\,2^{K+1}K(N+1)^{K-1})$ operations per procedure $i$, which is quadratic in $N$ for two procedures. For the small menus in our experiments, the implementation also solves the equivalent multiple-choice integer program, grouping items with identical costs and reduction vectors.

\paragraph{Hardness.}
Consider an unweighted set-cover instance with $\nu$ elements, each contained in at least one set. Create an original that answers A everywhere and one rival per element. For every set, create one item whose key is hidden, costs one, and equals A. Rival $j$ answers B on this item exactly when element $j$ belongs to the set, and A otherwise. All answers are available. Let $f_j$ be the number of sets containing $j$. Add $\max_j(f_j-1)$ fully revealed padding items with key A, on exactly $f_j-1$ of which rival $j$ answers B. Rival $j$'s initial largest lead over the original is then $f_j-(f_j-1)=1$. Opening the key of a set that contains $j$ turns its possible $+1$ on that item into a realized $-1$, a reduction of two; other keys change nothing for $j$. On the completed record the original is the unique best procedure. With zero tolerance, a set of opened keys therefore certifies the original exactly when the corresponding sets cover all $\nu=K-1$ elements. The optimal costs coincide, and feasible solutions map back without increasing cost. The logarithmic inapproximability of set cover \cite{sethard} transfers. Combined with $H_{2(K-1)}\le 1+\ln(2(K-1))$, this matches the leading logarithmic dependence on the number of rivals.

\section{Two limits of the decision model}\label{app:scope}
Two boundary cases help interpret a review protocol. A team may accept a proposed correction only some of the time, in which case the effect on accuracy depends on which changes it accepts. Alternatively, the team may know every correct answer yet still lack some of the reviews it needs to compare two procedures. Each case identifies a different piece of information the decision requires.

\paragraph{Partial adoption at the population level.}
Let $e$ and $r$ be the error probabilities of the producer's and the reviewer's answers under the specified protocol, and let $\pi$ be the probability that both are wrong. Suppose the released answer adopts the reviewer's answer with probability $\alpha$ when this would repair an error and with probability $\beta$ when it would spoil a correct original. Counting the two cells in which exactly one answer is correct, the expected gain in accuracy is
\[
  \alpha(e-\pi)-\beta(r-\pi).
\]
From the four-cell probability table, $\pi$ can take any value in $[\max(0,e+r-1),\min(e,r)]$. With one producer and several reviewers, these ranges can be attained jointly by drawing each reviewer's error conditionally on the producer's. A rate conditioned on a cell of probability zero has no effect. When the reviewer's answer is always adopted, the gain is $e-r$. Here $r$ is the error rate of the answer the reviewer gives \emph{after} reading the original, not of its independent answer. The 51-answer difference in the introduction is consistent with this identity.

\paragraph{When a procedure's answers must also be bought.}
Suppose all $N$ keys are known and there are two procedures. On items where both answers are available, let the partly observed procedure lead by $g$ correct answers. On items where its answer is missing, let the fully observed procedure be correct $c$ times and wrong $w$ times. If each missing answer can independently be right or wrong, the total difference can be any integer from $g-c$ to $g+w$: making every missing answer wrong attains the lower end, and making one more of them right raises the lead by exactly one. Buying keys cannot narrow this interval. For example, with $N=100$, $g=8$, $c=12$ and $w=8$, the lead ranges from $-4$ to $16$, and at least three favourable answers must be revealed to certify the partly observed procedure within one answer of its rival. This is the case in which caching every answer, as assumed in Section~\ref{sec:keys}, has a real cost.

\section{Implementation and evaluation details}\label{app:replay}
Replaying an audit requires a precise rule for breaking ties, converting a percentage tolerance into an integer deficit, and charging each query. These details determine when two runs using the same answers can return different choices or costs. We give the conventions used in the experiments, the entropy comparator, and the model-call accounting needed to reproduce the reported comparisons.

\paragraph{Analysis protocol.}
The menu order starts with keeping the producer's answer, followed by reviewers in lexical order of their names (MMLU) or in prompt order (numerical reports). Stand-alone and full ranking break ties by menu order. The tolerance is implemented as an integer number of answers (seven for 700 items, two for 256). In all figures and tables, every method uses the same precomputed answers within a menu. The offline minimum is solved separately for every feasible procedure, with identical coefficient rows grouped. The solver checks integer feasibility and agreement between the incumbent objective and its dual bound. Exhaustive tests enumerate all key completions for every subset of 20 small random pools, including invalid answers, and compare both the bounds and the minimum certificate sizes.

\paragraph{Model Selector comparator.}
Our NumPy implementation follows the default pool-based Model Selector, with $\eta=0.46$, in Kay et al.'s evaluation code. Starting from equal weights on the procedures, it multiplies the weights of procedures that answer an opened key correctly by $(1-\eta)/\eta$ and renormalizes. To score an item, it simulates each of the four possible keys and averages the entropy of the resulting weights uniformly. It opens the disagreement with the lowest expected entropy and opens agreements only when no disagreement remains. Invalid answers match none of the four keys. All acquisition methods share seeded tie priorities and our deterministic stopping rule. We did not tune $\eta$; the comparison concerns acquisition cost under a common stopping rule.

\paragraph{Data and computational cost.}
The 32,000 MMLU calls (8,000 independent answers and 24,000 reviews) were made in the earlier exposure study; this study made no new MMLU queries. The 24,000-row score table repeats each reviewer's independent answer across the three producers it reviewed. In Study~2, the 896 items required seven local calls each: 6,272 calls in total, with 627,795 prompt tokens and 12,544 completion tokens as reported by the runtime. All producer and reviewer answers were generated before the audit simulations. The primary comparator, splits and tolerance were recorded before the new answers were generated. The Model Selector comparison, the prior ablation, the tolerance sweep, the wider menu and the Singapore replay are supplementary analyses. MMLU independent answers were generated in the producer role, and the review prompts included a justification, which limits the causal interpretation of their difference. Model configuration, output grammar and data-generation seeds for Study~2 are in the release.

\end{document}